\documentclass{article}
\usepackage{log_2023}            

\usepackage{booktabs}            
\usepackage{multirow}            
\usepackage{amsfonts}            
\usepackage{graphicx}            
\usepackage{duckuments}          

\usepackage[utf8]{inputenc} 
\usepackage[T1]{fontenc}    
\usepackage{url}            
\usepackage{nicefrac}       
\usepackage{microtype}      
\usepackage{xcolor}         
\usepackage{bm}
\newcommand{\R}{\mathbb{R}}
\usepackage{algorithm}
\usepackage{algpseudocode}
\usepackage{times, amsfonts, amsmath, amssymb}
\usepackage{amsthm}
\usepackage{subcaption}
\usepackage[font=footnotesize]{caption}

\newcommand{\changed}[1]{\textcolor{black}{#1}}

\usepackage[numbers,compress,sort]{natbib}

\title[Intrinsically Motivated Graph Exploration Using Network Theories of Human Curiosity]{Intrinsically Motivated Graph Exploration Using Network Theories of Human Curiosity}

\author[Patankar et al.]{
Shubhankar P. Patankar$^{*}$ \quad Mathieu Ouellet$^{*}$ \quad Juan Cervi\~no \quad Alejandro Ribeiro \\
\textbf{Kieran A. Murphy} \quad \textbf{Dani S. Bassett} \\
University of Pennsylvania\\
\texttt{\{spatank,ouellet,jcervino,aribeiro,kieranm,dsb\}@seas.upenn.edu}
}

\begin{document}

\maketitle

\begin{abstract}
    Intrinsically motivated exploration has proven useful for reinforcement learning, even without additional extrinsic rewards.  
    When the environment is naturally represented as a graph, how to guide exploration best remains an open question.  
    In this work, we propose a novel approach for exploring graph-structured data motivated by two theories of human curiosity: the information gap theory and the compression progress theory.  
    The theories view curiosity as an intrinsic motivation to optimize for topological features of subgraphs induced by nodes visited in the environment.
    We use these proposed features as rewards for graph neural-network-based reinforcement learning.  
    On multiple classes of synthetically generated graphs, we find that trained agents generalize to longer exploratory walks and larger environments than are seen during training.
    Our method computes more efficiently than the greedy evaluation of the relevant topological properties.  
    The proposed intrinsic motivations bear particular relevance for recommender systems. 
    We demonstrate that next-node recommendations considering curiosity are more predictive of human choices than PageRank centrality in several real-world graph environments.
\end{abstract}

\section{Introduction} \label{sec:introduction}
\def\thefootnote{*}\footnotetext{Equal contribution. Code available at: \url{https://github.com/spatank/curiosity-graphs}.}

Providing a task-agnostic incentive for exploration as an intrinsic reward has proven useful for reinforcement learning, even in the absence of any task-specific (extrinsic) rewards~\cite{Pathak_2017,Burda_2018largescale}.
Termed \textit{curiosity} in reference to the analogous drive in humans, prior formulations are based on different means of quantifying the novelty or surprisal of states encountered by an agent~\cite{Guo_2022byolexplore}.
If states are represented as graphs, the task-agnostic motivation to explore can additionally be content-agnostic, depending only on the topological properties of the visited state subgraph.
Leading theories of human curiosity are similarly content-agnostic, based only on the structural properties of a relational graph connecting atoms of knowledge without regard to their actual content \cite{Zhou_2020}.

Theories of curiosity seek to describe the intrinsic motivations that underlie human decision-making when acquiring information through exploration.
The \textit{information gap theory} (IGT) argues that curiosity collects knowledge to regulate gaps in our understanding of the world \citep{Loewenstein_1994}.
Exposure to a small amount of novel information pushes an individual's uncertainty about the environment past an acceptable threshold, creating an information gap.
Curious agents are driven to resolve the discrepancy by acquiring information to close the gap \citep{Kang_2009, Daddaoua_2016}.
An alternative account, the \textit{compression progress theory} (CPT), posits that information-seeking behavior is motivated to build increasingly compressible state representations \citep{Schmidhuber_2008, Schmidhuber_2010}.
Compression enables abstraction and improved generalization by emphasizing the essential latent structures of knowledge \citep{Tenenbaum_2011, Collins_2017, momennejad2020learning}.
Both theories provide optimization objectives for the human exploration of graph-structured environments.

\begin{figure}[htbp] 
    \begin{center}
    \includegraphics[width=\textwidth]{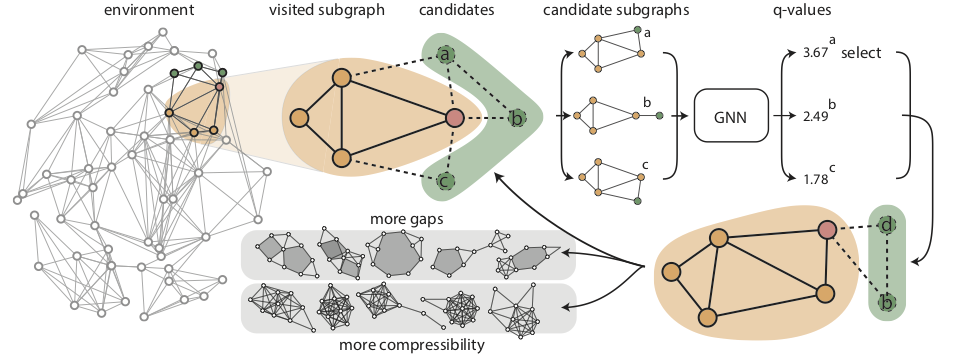}
    \end{center}
    \caption{\label{fig:schematic_1} \textbf{Neural network for graph exploration.} The subgraph induced by the set of currently visited nodes is denoted in orange. Candidate nodes to visit at the next time step are denoted in green. We build candidate subgraphs by adding each neighbor to the already visited subgraph. The candidates are processed with a GNN to obtain Q-values, denoting their long-term potential to create or close gaps or to improve compressibility. Two example trajectories are shown: one with a high number of gaps and one with greater compressibility.}
\end{figure}


In this work, we demonstrate that network theoretic measurements of information gaps and compression progress can be meaningful exploration incentives for graph neural network (GNN)-based reinforcement learning (RL).
\changed{Here, similar to human curiosity, which is typically conceived of as being non-instrumental, and unlike traditional RL formulations, where artificial curiosity is a means to an end, exploration itself is the broader goal.}
To that end, we train GNN agents to explore graph-structured environments while optimizing for gap creation and improved compression (Figure \ref{fig:schematic_1}). 
\changed{Additionally, we use GNNs trained with human curiosity rewards to modify PageRank centrality.
We measure an alternative form of PageRank by biasing underlying random walks towards nodes that create information gaps or improve network compressibility.
We use human exploration trajectories acquired from spaces that can be naturally represented as graphs---movies, books, and Wikipedia---to evaluate predictions of user choices made by PageRank against those made by our new metric.}

Our primary contributions are the following:
{
\setlength{\leftmargini}{0.5cm}  
\begin{itemize}
  \setlength{\itemsep}{1pt}
  \setlength{\parskip}{0pt}
  \setlength{\parsep}{0pt}
    \item We adapt intrinsic motivations for human curiosity as reward functions for reinforcement learning.
    \item We replace expensive reward computations with graph neural networks. \changed{Subject to training costs,} our method is computationally efficient and generalizes to longer exploratory walks and larger environments than are seen during training.
    \item We demonstrate that incorporating curiosity into PageRank centrality leads to better predictions of human preferences compared to standard PageRank.
\end{itemize}
}


\section{Related work}
\label{sec:related}
\textbf{Human curiosity as graph exploration.} Curiosity in humans is conceptualized as the intrinsic motivation to gather information from the environment \citep{Loewenstein_1994, Gottlieb_2013, Kidd_2015}.
Humans acquire information even when it is expensive \citep{hsee2016pandora, clark2021smokers} and may have no tangible utility \citep{Bennett_2016, Brydevall_2018}, suggesting that exploration is inherently valuable. 
\changed{Recent work has expanded the traditional knowledge acquisition perspective on curiosity by also considering how units of knowledge relate to each other.}
This perspective defines curiosity as an exploratory walk on a graph.
Here, curiosity entails building a growing knowledge network by acquiring informational units as nodes and their relationships as edges \citep{zurn_bassett_2018, Zhou_2020}. 
The state of an individual's knowledge is viewed as the subgraph of the environment induced by the visited nodes \citep{LydonStaley_2021, patankar2022curiosity}.
Under this formulation, humans explore Wikipedia via trajectories with fewer information gaps and greater network compressibility than relevant null models \citep{patankar2022curiosity}.

\textbf{Intrinsic motivations in reinforcement learning.} 
The need for improved exploration has led reinforcement learning to incorporate curiosity-like intrinsic motivations into its algorithmic framework \cite{Aubret_2019, Ladosz_2022}.
Exploration rewards in RL take several forms.
At the core of all approaches is an inducement for the learning agent to seek novelty.
Count-based approaches encourage visits to unfamiliar or infrequently visited states \cite{Bellemare_2016, Ostrovski_2017, Tang_2017, Machado_2018, Jo_2022}.
When the state space is large, enumerating the frequencies of visits to all possible states is expensive.
To overcome this challenge, density models derive uncertainty-based pseudo-counts \citep{Bellemare_2016, Ostrovski_2017}.
A complementary perspective emphasizes model building and formulates curiosity in terms of learning progress and surprisal \cite{Oudeyer_2007_1, Oudeyer_2007_2, Schmidhuber_2010, Stadie_2015, Houthooft_2016, Pathak_2017}. 
For instance, in the prediction error approach---alongside an extrinsic task---the agent attempts to learn a model of the environment's dynamics.
Curiosity rewards are proportional to the error when predicting transitions between states.
Memory-based methods assign rewards considering how different a newly visited state is from those stored in memory \cite{Fu_2017, Savinov_2018}.
Instead of a prescriptive approach, parametric methods attempt to explicitly learn an intrinsic reward function \cite{Song_2010, Zheng_2018, Memarian_2021, Devidze_2022}. 
In general, improved exploration is a means to an end, with intrinsic rewards supplementing extrinsic task-specific rewards.

\textbf{Graph combinatorial optimization and reinforcement learning.} Combinatorial optimization entails selecting elements from a finite set of options such that the chosen subset satisfies an objective function \cite{Papadimitriou_1982}. 
Graph analyses often involve combinatorial optimization, with graph structure imposing constraints on the solution space. 
Recent work combines graph neural networks and reinforcement learning to construct solutions by incrementally adding nodes to a partial set \cite{Peng_2021, Mazyavkina_2021, Munikoti_2022_2}.
First, a GNN constructs an embedding for the candidate solution; second, an agent, for instance, a deep Q-network (DQN), trained via RL, selects an action to expand the solution \cite{Dai_2017}.
The two networks can be trained end-to-end with an optimization objective driving gradients for learning.
This approach solves various graph combinatorial tasks, such as the traveling salesperson problem \citep{Dai_2017, Joshi_2020, Hu_2020}, finding the maximum independent set \cite{Ahn_2020}, or the minimum vertex cover \cite{Dai_2017, Khzam_2022}, and identifying isomorphic subgraphs \cite{Wang_2022}. 
Instead of uncovering nodes, GNNs can also sequentially collapse nodes into each other with implications for matrix multiplication \cite{Meirom_2022}. 
\changed{Recent work has also sought to formulate the graph exploration task explicitly as a Markov decision process, using domain-specific node features and novelty rewards \cite{Dai_2019, Chen_2021}.}
GNNs, in combination with RL, have also been used to build and rewire graphs such that they possess high values of specific features of interest \cite{Darvariu_2021, Doorman_2022}.
\changed{}

\textbf{PageRank and human navigation on graphs.} PageRank seeks to model and predict online human browsing preferences.
PageRank assigns centrality scores to web pages, considering their importance and relevance, determined by the number and quality of links between pages \cite{gleich2015pagerank}.
It employs a random walks-based approach that includes occasional jumps, known as teleportation, to simulate the likelihood that users will transition between different pages \cite{zhu2022identification}.
Beyond its initial application to serving search results, PageRank has found broad utility in modeling human navigation in other graph-structured environments \cite{gleich2015pagerank}.
PageRank-based recommendations show alignment with empirical observations of human behavior \cite{ficzere2022random,piccardi2023large,aguinaga2015concept,benotman2021comparing}.
Further, topological characteristics of the underlying graph impact user navigation, underscoring the importance of considering connectivity patterns when forecasting noisy human preferences \cite{dimitrov2017makes}.
The PageRank algorithm explicitly factors topology into its computations by adjusting the underlying random walk process, aligning either the transition or teleportation weights with the environment’s topological features \cite{espin2019hoprank,piccardi2023large}.

\section{Methods}
Our goal is to train an agent to explore while optimizing for a structural property of the visited subgraph.
Consider a graph-structured environment $\mathcal{G} = (\mathcal{V}, \mathcal{E})$ with node set $\mathcal{V}$ and edge set $\mathcal{E} \subseteq \mathcal{V} \times \mathcal{V}$. 
Let $\mathcal{V}_T = \{v_1, v_2, \cdots, v_{T}\} \subseteq \mathcal{V}$ be an ordered set of explored nodes at time $T$.
The corresponding subgraph trajectory is the sequence $\mathcal{S}_{1} \subset \mathcal{S}_{2} \subset \cdots \subset \mathcal{S}_{T}$, wherein the $t$-th subgraph $\mathcal{S}_t$ is induced by the first $t$ visited nodes.
Specifically, given the graph $\mathcal{G}$, the number of nodes to visit $T$, a graph feature function $\mathcal{F}: 2^{\mathcal{G}} \rightarrow \mathbb{R}$, and a discount factor $\gamma \in [0,1]$, we seek an ordered set $\mathcal{V}^*_T$ such that $\sum_{t=1}^{T}\gamma^{t-1}\mathcal{F}(\mathcal{S}_t)$ is maximal.
The function $\mathcal{F}$ acts as an intrinsic reward to encourage exploration.
The discounting parameter determines the extent to which future values of $\mathcal{F}$ factor into the decision-making at each step.
Drawing inspiration from human curiosity, we adopt information gap theory and compression progress theory to design two \changed{reward} functions, $\mathcal{F}_{IGT}$ and $\mathcal{F}_{CPT}$.

\subsection{Network theories of curiosity}
\label{sec:curiosity}

\begin{figure}[htbp] 
    \begin{center}
    \includegraphics[width=\textwidth]{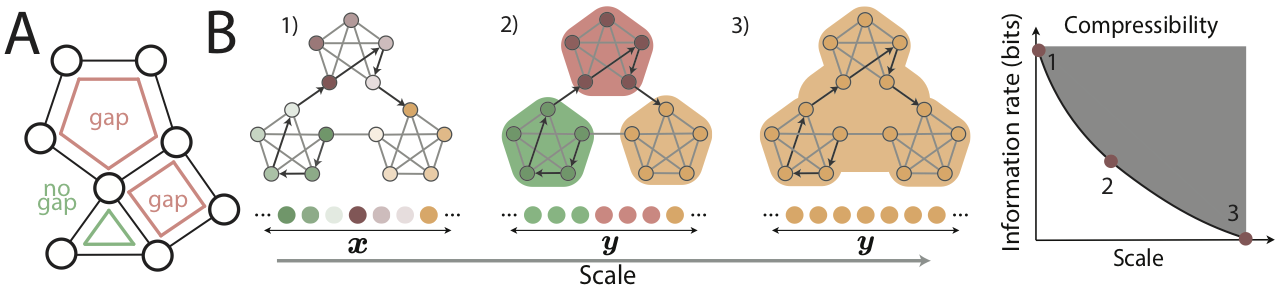}
    \end{center}
    \caption{\label{fig:schematic_2} \textbf{Quantifying network theories of human curiosity.} \emph{(A)} Gaps or cavities in a graph can be formalized using algebraic topology. A $1$-dimensional cavity, also known as a 1-cycle, is a non-triangular loop of edges. \emph{(B)} The information rate of a random walk $\boldsymbol{x}$ on a graph is given by its entropy. If we cluster the nodes, the walk sequence $\bm{x}$ is compressed into a new sequence $\boldsymbol{y}$, where $y$ is the cluster that contains node $x$. The new sequence has a lower information rate than the original sequence. The number of clusters defines the scale at which the network is described. We can find an optimal clustering at every scale of description that maximally lowers the information rate. These values can be recorded in a rate-distortion curve. Network compressibility is the maximal reduction in the information rate, averaged across all scales. Graphically, this value represents the area above the rate-distortion curve bounded by the entropy of the unclustered random walk.}
\end{figure}

\textbf{Information gap theory} views human curiosity as an intrinsic motivation to regulate gaps in knowledge. 
Exposure to new information pushes the level of uncertainty about the environment past an acceptable threshold, creating an uncertainty gap.
Curiosity seeks to find information units to close this gap.
By modeling the state of knowledge as a graph, we can characterize information gaps as topological cavities. 
In a graph, cavities can take several forms: dimension $0$ cavities represent disconnected network components, whereas those of dimension $1$, known as $1$-cycles, represent non-triangular loops of edges (Figure \ref{fig:schematic_2}A). 
In order to identify and count topological cavities, a graph is first converted into a higher-order relational object known as a \textit{simplicial complex} \cite{Gross_1987}.
A simplicial complex is comprised of simplices.
Geometrically, a $d$-simplex is a shape with flat sides formed by connecting $d+1$ points.
For $0 \leq d \leq 2$, by definition a node is a $0$-simplex, an edge is a $1$-simplex, and a filled triangle is a 2-simplex.
We can construct a simplicial complex by assigning a $d$-simplex to each $(d+1)$-clique in a binary graph.
In a simplicial complex, a $d$-dimensional topological cavity is identified as an enclosure formed by $d$-simplices that cannot be filled by a higher-dimensional simplex.
We refer the reader to Refs. \citep{Carlsson_2009, Ghrist_2007, Hatcher_2002, Zomorodian_2005, Bianconi_2021} for more details on algebraic topology.

Given a simplicial complex, the $d$-th \textit{Betti number} $\beta_d$ counts the number of topological gaps of dimension $d$. 
Prior work examining human curiosity finds compelling evidence in support of information gap theory.
\changed{In particular, humans create induced subgraphs with an increasing number of $1$-dimensional cavities \citep{patankar2022curiosity}.}
Therefore, in this work, at each time step $t$ with a visited subgraph $\mathcal{S}_t$, we assign rewards equal to $\beta_1$, that is, we set $\mathcal{F}_{IGT} = \beta_1(\mathcal{S}_t)$.


\textbf{Compression progress theory} posits that curiosity is a drive to compress the state of knowledge~\cite{Schmidhuber_2008}.
During graph exploration, at each step $t$ in a trajectory, a reward for compression can be assigned as network compressibility \citep{Lynn_2021}.
Consider a subgraph $\mathcal{S}_t$ with $t$ nodes and $q$ edges, represented by a symmetric adjacency matrix $M \in \R^{t \times t}$. 
Information about the subgraph's structure can be encoded in the form of a random walk $\boldsymbol{x} = \left(x_{1},\; x_{2},\; \ldots\;\right)$. 
The walk sequence is generated by randomly transitioning from a node to one of its neighbors. 
Thus, for a random walk on $\mathcal{S}_t$, the probability of transitioning from node $i$ to node $j$ is $P_{ij} = M_{ij}/\sum_jM_{ij}$. 
Since the walk is Markovian, its information content (or \textit{entropy}) is given by $H = -\sum_{i} \pi_{i} \sum_{j} P_{i j} \log P_{i j}$.
Here, $\pi_i$ is the stationary distribution representing the long-term probability that a walk arrives at node $i$, given by $\pi_i = \sum_jM_{ij}/2q$.

Assigning nodes to clusters leads to a coarse-grained sequence $\boldsymbol{y} = \left(y_{1},\; y_{2},\; \ldots\;\right)$. 
The number of clusters $n$ can be used to define a scale of the network's description $s = 1 - \frac{n-1}{t}$. 
When $n = t$, the network is described at a fine-grained scale $s = 1/t$; at the other extreme, when $n = 1$ the network is described at the coarsest scale $s = 1$. 
At every description scale in between, it is possible to identify a clustering of nodes that minimizes the information rate (Figure \ref{fig:schematic_2}B). 
After computing these optimal clusterings across all scales, we arrive at a rate-distortion curve $R(s)$, representing a bound on the information rate as a function of the scale $s$. 
The compressibility $C$ of the network is then given as the average reduction in the information rate across all scales \citep{Lynn_2021}, $C = H - \frac{1}{t}\sum_{s} R(s)$.
\changed{Human curiosity in graph-structured environments leads to induced subgraphs with increasing network compressibility \cite{patankar2022curiosity}.}
Therefore, we assign compression rewards as $\mathcal{F}_{CPT} = C(\mathcal{S}_t)$,
where $C(\mathcal{S}_t)$ denotes the compressibility of subgraph $\mathcal{S}_t$.

\subsection{Reinforcement learning for graph exploration}
\label{sec:MDP}


We formulate the graph exploration problem as a Markov decision process (MDP) \citep{Sutton_2018}: 
{
\setlength{\leftmargini}{0.5cm}  
\begin{itemize}
  \setlength{\itemsep}{1pt}
  \setlength{\parskip}{0pt}
  \setlength{\parsep}{0pt}
    \item \textbf{States}: The state is defined as the subgraph induced by the visited nodes at time $t$, $\mathcal{S}_t = \mathcal{G}[\mathcal{V}_t]$. We specify the initial state $\mathcal{S}_1$ by randomly selecting a starting node $v_1 \in \mathcal{V}$. Each state represents a partial solution to the broader sequential exploration task. 
    \item \textbf{Actions}: The agent can transition to a neighbor of the most recently visited node. 
    We denote the neighborhood of a node $v$ as $\mathcal{N}(v) = \{u \in \mathcal{V} \mid(v, u) \in \mathcal{E}\}$. Therefore, given the state at time $t$, the set of available next nodes is $\mathcal{A}(\mathcal{S}_t) = \mathcal{N}(v_t)\backslash\mathcal{V}_t$. 
    If no nodes are available in the immediate neighborhood, we expand the action set to include all neighbors of the explored subgraph. 
    \item \textbf{Transitions}: Given the pair $\mathcal{S}_t$ and $v \in \mathcal{A}(\mathcal{S}_t)$, the transition to state $\mathcal{S}_{t+1}$ is deterministic with $P\left(S_{t+1} \mid S_{t}, v\right) = 1$.
    \item \textbf{Rewards}: The reward at time $t$ is defined as $R_t = \mathcal{F}(\mathcal{S}_t)$.
    \changed{Considering information gap theory, we reward agents for visiting nodes that create 1-cycles ($\mathcal{F}_{IGT}$). Considering compression progress theory, we reward agents for visiting nodes that improve network compressibility ($\mathcal{F}_{CPT}$).} 
\end{itemize}
}
The policy $\pi(v \mid \mathcal{S}_t)$ maps states to actions, fully describing the agent's behavior in the environment. 
At each step, the agent makes decisions using a value function $Q(\mathcal{S}_t, v)$, which evaluates candidate nodes $v \in \mathcal{A}(\mathcal{S}_t)$ in the context of the currently explored subgraph $\mathcal{S}_t$. 
The function measures the total (discounted) reward that is expected to accumulate if the agent selects action $v$ in state $\mathcal{S}_t$ and thereafter follows policy $\pi$.
In turn, the policy can be viewed as behaving greedily with respect to the value function, $\pi = \arg \max _{v \in \mathcal{A}(\mathcal{S}_t)} Q\left(\mathcal{S}_t, v\right)$.
Solving an MDP entails finding an optimal policy that maximizes the expected discounted sum of rewards.

We parameterize the value function $Q$ using a GNN $\Phi(\cdot):\mathcal{G}\to\mathbb{R}$.
GNNs build vector embeddings for nodes by iteratively aggregating their features with those from their local neighborhoods \cite{Zonghan_2021}.
Each aggregation step is typically followed by a fully connected layer and a non-linear activation function.
Depending on the number of rounds of aggregation, features from more distant locations in the graph can inform the embedding for each node.
Specifically, we use the \textit{GraphSAGE} architecture \cite{Hamilton_2017}, where at the $l$-th round of feature aggregation, the embedding for node $u$ is given as,
\begin{equation}
    h_u^{(l)}=f^{(l)}\left(h_u^{(l-1)}, h_{\mathcal{N}(u)}^{(l-1)}\right)=g\left[\theta_C^{(l)} h_u^{(l-1)}+\theta_A^{(l)} \tilde{A}\left(h_{\mathcal{N}(u)}^{(l-1)}\right)\right],
\end{equation}
where $\tilde{A}$ represents the aggregation operator, $g\left[.\right]$ is the activation function, and $\theta_C$ and $\theta_A$ are parameters for combination and aggregation, respectively \cite{Hamilton_2017, Munikoti_2022_2}. 
\changed{Our choice of GraphSAGE is motivated not by its sampling-based approach but rather by its capacity for inductive learning. 
Therefore, we do not perform neighbor sampling during feature aggregation.
Further, considering that leading network theories of human curiosity are agnostic to the precise content of individual nodes, we only use the local degree profile (LDP) of each node as the initial feature set \cite{Cai_2018}.}
LDP comprises features of a node's neighborhood, including its degree, the minimum and maximum degrees of its neighbors, and the average and standard deviation of the degrees of its neighbors.

We train GNNs for exploration using the DQN algorithm, with a replay buffer for experience sampling, a target network, and a decaying $\epsilon$-greedy exploration rate \citep{Mnih_2015}.
Details of the full neural network architecture and the training process are included in the Supplement.




\subsection{Curiosity-biased node centrality}
Several graph theoretical quantities can be defined in terms of random walk processes.
We can use agents trained to explore graphs to bias random walkers and, by extension, the corresponding quantities.
PageRank is a widely recognized algorithm that assigns node centrality scores to graph data \cite{brin1998anatomy,wu2007extracting,gleich2015pagerank,zhu2022identification}.
The per-node score $\eta$ can be interpreted as the stationary distribution of a random walk process on a network.
With probability $\alpha$, a random walker moves along an edge from node $v_i$ to one of its neighbors.
The probability of reaching a connected node $v_j$ is $P_{ij}$.
Alternatively, with probability $1-\alpha$, the walker jumps, or teleports, to a random node in the network.
The probability of jumping to node $v_k$ is $q_k$. 
Under conditions of irreducibility and aperiodicity \cite{haggstrom2002finite}, the stationary distribution is given as 
\begin{align}
\sum_i (I-\alpha P_{ij}^t)\eta_i = (1-\alpha)q_j.
\end{align} 
The PageRank algorithm follows a random walk that is entirely Markovian. 
Typically, the probability $P_{ij}$ depends solely on the out-degree of node $v_i$ and, in the case of node-weighting, on the vector $q$.
Personalized PageRank biases the random walk process using $q_k$ by taking into account nodes that are already visited in the network~\cite{haveliwala2003analytical}. 

\changed{Adaptations of PageRank often incorporate biases in the weighting scheme or the teleportation mechanism to better predict human navigation in graph-structured environments \cite{ficzere2022random,espin2019hoprank,dimitrov2017makes,piccardi2023large,aguinaga2015concept,benotman2021comparing}.}
We can integrate agents trained to optimize for the exploration objectives described earlier into the PageRank algorithm in the form of similar biases. 
Specifically, given an already visited subgraph, we propose to modify transition probabilities using the Q-values assigned to candidate nodes. 
Consider a non-Markovian random walker sitting at node $v_l$ with a path history $V_l= \{ v_1, \cdots,v_{l-1},v_{l} \}$.
The visited nodes in the path induce a corresponding subgraph $\mathcal{S}_l$.
Paths are built starting from the most recent initialization or teleportation event.
We use a Q-value function trained to optimize for an objective $\mathcal{F}$ to bias the walker.
The transition probability from node $v_l$ to node $v_m$ can be re-defined as,
\begin{align}
P^{\mathcal{F}}_{lm}(\mathcal{S}_l) \equiv \begin{cases}
    \frac{(1-p_g)p_g^{rank(Q(\mathcal{S}_l, v_m))-1}}{1-p^{| \mathcal{A}(\mathcal{S}_l)|}}, & v_m \in \mathcal{A}(\mathcal{S}_l),\\
    0, & \text{otherwise},
  \end{cases} 
\end{align}
where $rank(Q(\mathcal{S}_l, v_m))$ is the rank for $v_m$ considering the Q-values for the candidate nodes and $p_g\in [ 0,1]$ is a parameter that controls how likely the walker is to select nodes greedily. 
To compute biased per-node PageRank values, we simulate a walker using $P^{\mathcal{F}}_{ij}(\mathcal{S}_i)$ until probabilities converge. 


\section{Experiments}
\label{sec:Experiments}
\subsection{Exploration in synthetically generated networks}

We train a curiosity-based GNN agent to explore synthetically generated graph environments that exhibit a broad range of degree profiles and topologies \cite{Newman_2010, dall2002random}. 
We examine synthetic networks generated using the random geometric (RG), Watts-Strogatz (WS), Barab\'asi-Albert (BA), and Erd\"os-R\'enyi (ER) graph models.
Details surrounding the generation process are available in the Supplement.
For each of the four graph models, we build $100$ training, $10$ validation, and $10$ testing environments.
Each environment is constructed to have $N = 50$ nodes. 
Each episode lasts for $10$ steps and, therefore, consists of visits to $10$ distinct nodes.
After training, we evaluate the GNN agent in the testing environments against four baseline approaches:
{
\setlength{\leftmargini}{0.5cm}  
\begin{itemize}
  \setlength{\itemsep}{1pt}
  \setlength{\parskip}{0pt}
  \setlength{\parsep}{0pt}
    \item Random: Select a candidate node at random.
    \item Greedy: For each candidate node, build a candidate state subgraph. Evaluate the reward function for each subgraph and select the node that results in the biggest one-step improvement.
    \item Max Degree: Select the candidate node with the largest degree.
    \item Min Degree: Select the candidate node with the smallest degree.
\end{itemize}
}

\begin{table}[H]
\centering
\begin{tabular}{@{}lllllll@{}}
\toprule
$\mathcal{F}$ & $\mathcal{G}$ & Random               & Max Degree           & Min Degree           & Greedy                        & GNN                            \\ \midrule
IGT           & RG            & $0.312_{\pm{0.034}}$ & $0.010_{\pm{0.007}}$ & $0.144_{\pm{0.027}}$ & $1.495_{\pm{0.079}}$          & $\mathbf{2.308_{\pm{0.092}}}$  \\
              & WS            & $1.141_{\pm{0.068}}$ & $1.048_{\pm{0.068}}$ & $1.586_{\pm{0.082}}$ & $2.707_{\pm{0.103}}$          & $\mathbf{3.303_{\pm{0.106}}}$  \\
              & BA            & $7.593_{\pm{0.145}}$ & $2.565_{\pm{0.083}}$ & $3.932_{\pm{0.115}}$ & $19.332_{\pm{0.206}}$         & $\mathbf{21.970_{\pm{0.169}}}$ \\
              & ER            & $9.197_{\pm{0.144}}$ & $9.638_{\pm{0.162}}$ & $4.953_{\pm{0.127}}$ & $\mathbf{25.20_{\pm{0.164}}}$ & $24.058_{\pm{0.183}}$          \\ \midrule
CPT           & RG            & $8.607_{\pm{0.027}}$ & $8.928_{\pm{0.027}}$ & $7.864_{\pm{0.033}}$  & $\mathbf{9.615_{\pm{0.014}}}$  & $9.271_{\pm{0.017}}$            \\
              & WS            & $7.117_{\pm{0.021}}$ & $6.788_{\pm{0.025}}$ & $6.937_{\pm{0.021}}$ & $\mathbf{7.668_{\pm{0.012}}}$ & $7.174_{\pm{0.014}}$            \\
              & BA            & $6.926_{\pm{0.023}}$ & $8.526_{\pm{0.015}}$ & $5.899_{\pm{0.016}}$ & $\mathbf{8.669_{\pm{0.016}}}$ & $8.556_{\pm{0.010}}$            \\
              & ER            & $6.767_{\pm{0.020}}$ & $6.931_{\pm{0.019}}$ & $6.022_{\pm{0.017}}$ & $\mathbf{8.262_{\pm{0.016}}}$ & $7.880_{\pm{0.015}}$            \\ \bottomrule
\end{tabular} 
\caption{Performance of GNN-based agents using information gap theory (IGT) and compression progress theory (CPT) compared to four baseline methods (random, max degree, min degree, greedy). We compare results using the total average return gathered by agents in four types of synthetic graph environments (random geometric - RG, Watts-Strogatz - WS, Barabási-Albert - BA, Erdős-Rényi - ER).}\label{table:synthetic}
\end{table}

The total average reward gathered by the different agents is presented in Table \ref{table:synthetic}.
For the IGT reward, in all graph models except for ER, the GNN outperforms the greedy agent. 
By contrast, the one-step-ahead greedy agent consistently performs best for CPT, with the GNN a close second.
Baseline approaches broadly perform well compared to the GNN for CPT than they do for IGT.
When exploring a graph with the IGT objective, adding a single node can close several topological gaps simultaneously, requiring careful consideration of options. 
By contrast, compressibility is less sensitive to the choice of node at each step due to its strong correlation with the clustering coefficient \citep{Lynn_2021}.
If exploring inside a cluster, neighbors of a node are likely to be neighbors of each other, lowering the likelihood that a single choice will significantly alter long-term network compressibility.
For instance, the max degree baseline performs well for the CPT objective in random geometric graphs because high-degree nodes are centrally placed and surrounded by dense, highly clustered neighborhoods \cite{dall2002random}.
Barab\'asi-Albert graphs, similarly, have highly clustered cores due to preferential attachment in their generative process \cite{da2007exploring}.
Watts-Strogatz networks have high clustering when the edge rewiring probability is low.
As a result, even random exploration in such topologies tends to occur inside clusters leading to greater compressibility.
In support of this view, the minimum degree baseline, which is likely to select a node outside of a cluster, is typically further apart from the performance of the GNN compared to the other baselines.

\subsubsection{Trajectory length and environment size generalization}

\begin{figure}[htbp] 
    \begin{center}
    \includegraphics[width=\textwidth]{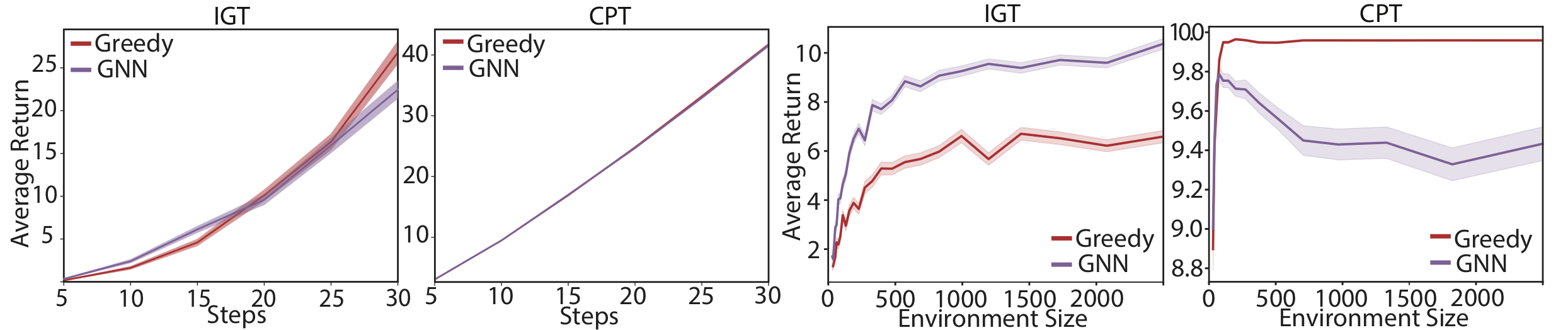}
    \end{center}
    \caption{\label{fig:fig_3} \textbf{Trajectory length and environment size generalization.} GNNs trained for graph exploration generalize to shorter and longer trajectories and to smaller and larger environments than are seen during training. We train GNNs to explore $10$ steps for IGT and CPT in random geometric environments with $50$ nodes. Performance does not degrade for exploratory walks of a different length in $50$-node environments. Similarly, when taking $10$ steps, GNN-based agents outperform or match the greedy agent in smaller and larger environments than those of size $50$ that are seen during training. Bands denote standard error.}
\end{figure}

After training the GNN agent to explore $10$ nodes in RG graph environments with $50$ nodes, we evaluate generalization performance for longer trajectories and larger environments.
We test trajectory length generalization while holding environment size fixed at $50$ nodes.
For walks shorter and longer than $10$ steps, the GNN performs comparably to the greedy agent for both IGT and CPT (Figure \ref{fig:fig_3}).
Next, we test environment size generalization by evaluating $10$-step walks in graph environments that are larger than $50$ nodes.
\changed{In environments that are up to two orders of magnitude larger than those seen during training, the GNN is consistently superior to the greedy agent for IGT and exhibits comparable performance for CPT (with an average return of 9.4 compared to 10).}
In summary, the performance of trained GNNs does not degrade for settings outside the training regime.
These results indicate that we can train GNNs for graph exploration in regimes where reward computations are relatively inexpensive due to the smaller size of subgraphs and expect them to scale to longer walks and larger networks.

\begin{figure}[htbp] 
    \begin{center}
    \includegraphics[width=0.65\textwidth]{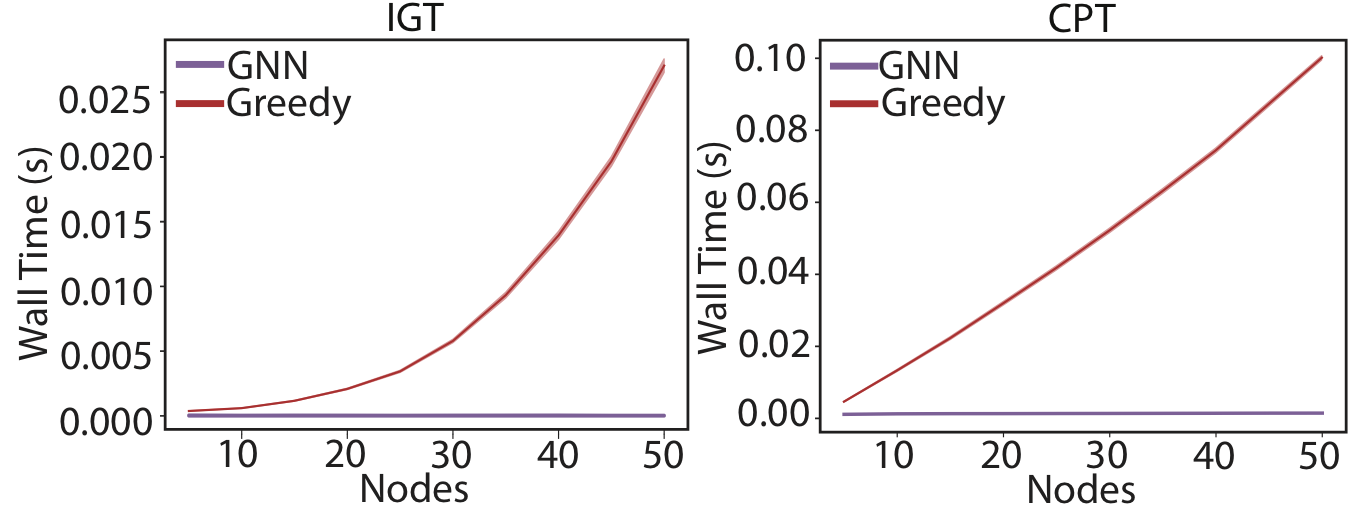}
    \end{center}
    \caption{\label{fig:fig_2} \textbf{Wall time.} Wall time for a forward pass through the GNN compared to the greedy evaluation of rewards. Bands denote standard error over computations for $50$ networks.}
\end{figure}

\subsubsection{Time complexity}
Using graphs of different sizes, we evaluate the computational efficiency of our approach by comparing the wall time for a forward pass through the GNN with that for a greedy evaluation of the two reward functions.
Figure \ref{fig:fig_2} displays results for RG graphs.
Wall time for greedy evaluation of the IGT and CPT objectives grows quickly with subgraph size, while the GNN offers a faster alternative.
\changed{Calls to $\mathcal{F}_{IGT}$ scale according to $O(|\mathcal{S}_t|^2)$ \cite{giunti2022average}, whereas those to $\mathcal{F}_{CPT}$ scale according to $O(|\mathcal{S}_t|^3)$.
By contrast, the computational complexity for the GNN grows linearly as $O(|\mathcal{S}_t|)$.}
Comparing the rewards for the two theories of curiosity, the information gap reward is significantly cheaper to evaluate compared to network compressibility.
Therefore, in addition to approximating human intrinsic motivations for exploration, we find that the GNN offers a route to efficient computation of meaningful topological features of graphs.

\subsection{Predicting human choices during graph navigation}

Next, we evaluate the utility of curiosity-trained agents for predicting human choices in graph-structured environments.
We gather human trajectories of graph exploration from three real-world datasets: MovieLens \cite{harper2015movielens}, Amazon Books \cite{he2016ups, mcauley2015image}, and Wikispeedia \cite{west2012human, west2009wikispeedia}.
Each dataset can be naturally represented as a graph-structured environment. 
Details on how we process the data are available in the Supplement.
We train GNNs for graph exploration in each environment for both information gap theory and compression progress theory.
We use GNNs trained with curiosity rewards to bias PageRank centrality to predict next nodes visited by humans.

To incorporate person-specific data when computing PageRank, we modify the hop vector $q$ to be zero for all nodes except a user's $n_\text{burn-in}$ most recently visited nodes~\cite{haveliwala2003analytical}. 
We assign a uniform jump probability to the $n_\text{burn-in}$ nodes, with $q_k = 1/n_\text{burn-in}$.
Each graph feature function $\mathcal{F}$ yields a PageRank vector $\eta^{\mathcal{F}}_i$. 
We combine these vectors linearly to obtain a final PageRank vector, denoted as $\eta'$ such that $\eta' \equiv \tilde{\beta} \eta_{\text{PR}}({\alpha}) + \tilde{\gamma} \eta_{\text{IGT}}({\alpha}) + \tilde{\delta} \eta_{\text{CPT}}({\alpha})$ where $ \tilde{\beta}^2 +\tilde{\gamma}^2 +\tilde{\delta}^2 = 1$ and $\eta_{\text{PR}}$ is the score vector obtained using standard PageRank. 
To evaluate this approach, we optimize the set of variables ${\alpha}, \tilde{\beta}, \tilde{\gamma}, \tilde{\delta}$ using a training set of transitions. 
We then compare performance against unbiased PageRank, where only ${\alpha}$ is optimized.
\changed{We split the set of user trajectories acquired from each dataset into $\mathbf{S}_{\text{test}}$ and $\mathbf{S}_{\text{train}}$.}
These sets consist of portions of human trajectories with a length of $n_\text{burn-in}+1$. 
Next, we perform Bayesian optimization to compute parameters $\hat{a}$ and $\hat{a}_\text{bias}$ for the two sets,
\begin{align}
\hat{a} &\equiv \arg\max_{\alpha} \sum_{S\in \mathbf{S}_{\text{train}}}\text{rank}_{v_{\text{burn-in}}}(\eta_{\text{PR}}(\alpha) )\\
\hat{a}_\text{bias} &\equiv \arg\max_{\alpha, \tilde{\beta}, \tilde{\gamma}, \tilde{\delta}} \sum_{S\in \mathbf{S}_{\text{train}}}\text{rank}_{v_{\text{burn-in}}}(\eta'(\alpha, \tilde{\beta}, \tilde{\gamma}, \tilde{\delta} ) ).
\end{align}
To evaluate our method, we calculate the ratio of improvement on the test set, given as 
\begin{align}
r_{\mathbf{S}_{\text{test}}} &\equiv \sum_{S\in \mathbf{S}{\text{test}}} \text{rank}_{v_{\text{burn-in}}}(\eta'(\hat{a}_\text{bias} ) ) / \sum_{S\in \mathbf{S}{\text{test}}} \text{rank}_{v_{\text{burn-in}}}(\eta_{\text{PR}}(\hat{a})  ) . 
\end{align}

Table \ref{table:pagerank} displays $r_{\mathbf{S}_{\text{test}}}$ in percentage terms for the three datasets when considering curiosity theories alone or in combination.
Across all combinations, improvement ranges from 2.9\% to 32.2\%, indicating that incorporating curiosity for the biasing of walks is useful.
The IGT or CPT-trained agents perform better with roughly similar values depending on the dataset.
In the Wikispeedia data, however, CPT leads to improvement nearly four times higher than IGT.
The books and movie datasets exhibit similarities since the selection mechanism in both environments is not directed towards a goal. 
By contrast, the Wikispeedia dataset involves goal-directed navigation.
\changed{We provide illustrative examples from the MovieLens dataset of user paths alongside predictions made by both unbiased and curiosity-biased PageRank in the Supplement.}

\begin{table}[H]
\centering
\begin{tabular}{@{}lccc@{}}
\toprule
Graph dataset $\mathcal{G}$ & IGT & CPT & IGT + CPT \\ \midrule
MovieLens   & $+4.2\%_{\pm{2.1\%}}$           & $+5.1\%_{\pm{1.5\%}}$            & $\mathbf{+7.9\%_{\pm{1.7\%}}}$     \\ \midrule
Amazon Books  & $+5.4\%_{\pm{1.9\%}}$           & $+4.6\%_{\pm{1.6\%}}$            & $\mathbf{+9.6\%_{\pm{1.9\%}}}$    \\ \midrule
Wikispeedia   & $+2.9\%_{\pm{2.9\%}}$           & $+12.2\%_{\pm{3.2\%}}$            & $\mathbf{+32.2\%_{\pm{7.7\%}}}$  \\ 
\bottomrule
\end{tabular}
\caption{Percentage improvement $(r_{\mathbf{S}_{\text{test}}})$ with curiosity-biased centrality for the MovieLens, Amazon Book, Wikispeedia datasets.}
\label{table:pagerank}
\vspace{-3mm}
\end{table}

\begin{figure}[htbp] 
    \begin{center}
    \includegraphics[width=\textwidth]{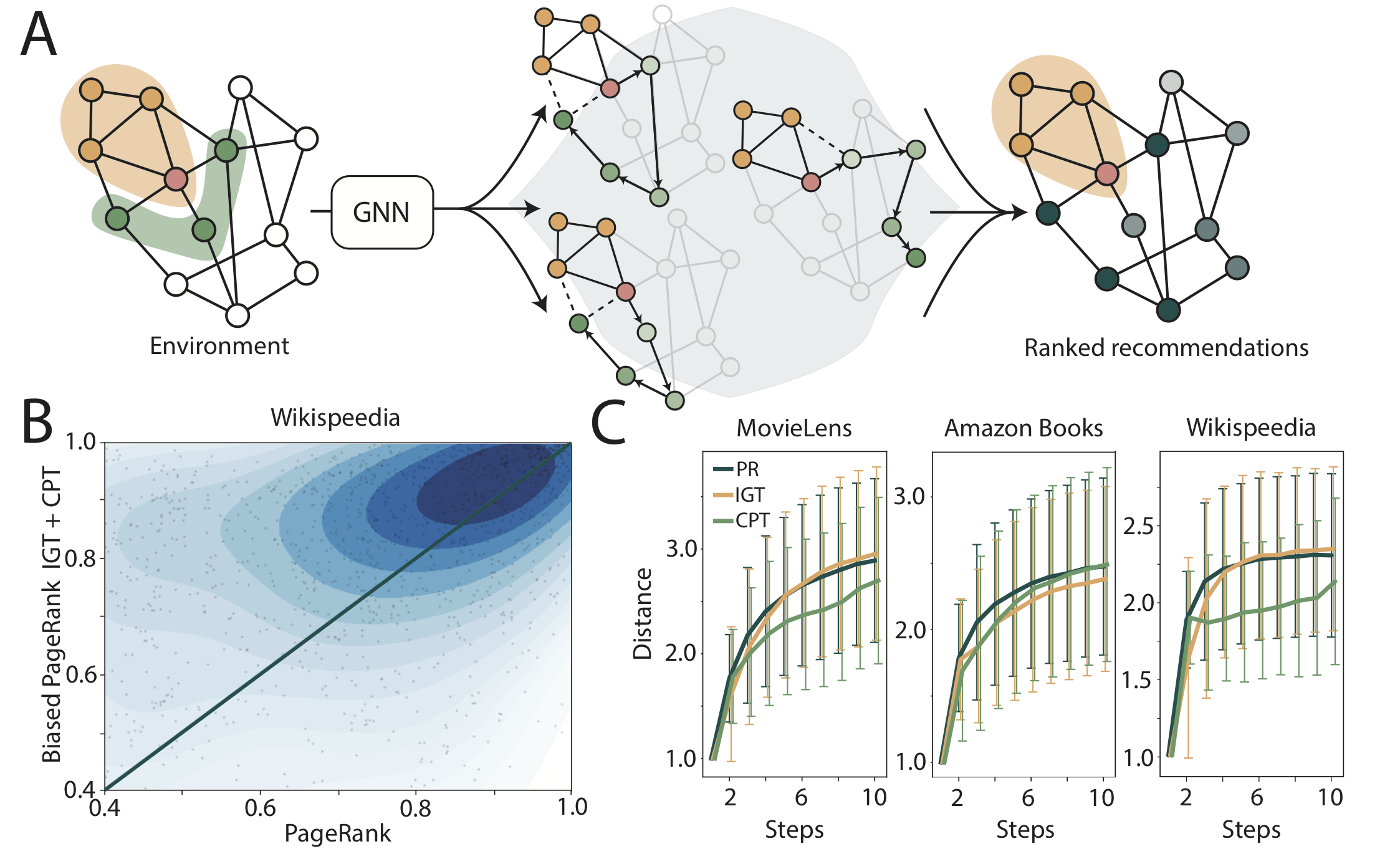}
    \end{center}
    \caption{\textbf{Re-defining centrality using agents trained for curiosity.} \emph{(A)} We measure curiosity-biased PageRank centrality using a set of biased walkers that explore the graph starting from a subset of already visited nodes. Biases are incorporated using GNNs trained for IGT and CPT rewards. 
    \emph{(B)} Example demonstrating the improvement in predicting human transitions when using curiosity-biased versus standard PageRank. Biased curiosity assigns higher percentile ranks to actual transitions than standard PageRank. \emph{(C)} Random walker diffusion, measured as the distance from the initial node for each graph. A comparison is made between the unbiased (blue), IGT-biased (orange), and CPT-biased walkers (green).}
    \label{fig:fig_4} 
\end{figure}

Figure \ref{fig:fig_4}B shows the improvement in predicting the transitions made by humans in the Wikispeedia dataset.
We compare percentile ranks for each transition made by the human when making predictions with and without biasing the random walk process.
We find that biased curiosity assigns higher percentile ranks to actual transitions than standard PageRank.
We also analyze the distance from the initial node with respect to time for individual random walk trajectories (Figure \ref{fig:fig_4}C).
In general, observed differences between the biased walkers are small and fall within the standard deviation of the walk process. 
However, the CPT-biased walker stands out as it tends to remain closer to the initial node in both the MovieLens and the Wikispeedia datasets \changed{(see Supplement)}.
These observations suggest that the differences observed in the biased PageRank algorithm are not solely attributable to changes in the diffusion properties of the random walks.

\section{Discussion}
\label{sec:Discussion}

\changed{How to measure curiosity best remains an open question both in the context of humans and for reinforcement learning.
While our specific choices---$1$-cycles and network compressibility---are motivated by recent work studying human behavior, other topological features may be more suited to drive graph exploration. 
Nonetheless, through our work, we demonstrate the utility of content-agnostic topology-aware intrinsic motivations.}
\changed{Similar to leading theories of human curiosity that are intentionally content independent, our method 1) uses no node information other than topological statistics and 2) uses no information in the reward other than what is available in the structure of visited subgraphs. 
Even at this level of abstraction, where topology takes precedence over content, we show that agents learn generalizable and transferable graph exploration strategies.
Further, we show that agents trained with human-like motivations can help devise centrality measures that predict human behavior better than PageRank. 
This result has two critical implications. 
First, we can use our method to test hypotheses about human motivations when navigating graph-structured environments. 
Trained agents can act as hypothesis testers by examining whether their choices for the subsequent nodes to visit align with human choices.
Second, our method can be used to design recommender systems for environments where human navigation of graphs is largely goalless.}


\bibliographystyle{unsrtnat}
\bibliography{references}

\end{document}